\documentclass[11pt,a4paper]{article}
\usepackage[hyperref]{emnlp2020}
\usepackage{times}
\usepackage{booktabs}
\usepackage{latexsym}
\usepackage{xspace}
\usepackage{graphicx}
\usepackage{amsmath}
\usepackage{tabularx}
\usepackage{multirow}

\usepackage{enumitem}

\newif\ifcomments
\commentstrue 
\ifcomments
    \providecommand{\matt}[1]{{\protect\color{teal}{\bf [Matt: #1]}}}

\else
    \providecommand{\matt}[1]{}
\fi

\usepackage{microtype}

\aclfinalcopy % Uncomment this line for the final submission
\newcommand\dataset{\textbf{IIRC}\xspace}

\newcommand\numquestions{$13441$}
\newcommand\numparagraphs{$5698$}

\title{IIRC: A Dataset of Incomplete Information\\Reading Comprehension Questions}

\author{James Ferguson$^\diamondsuit$\thanks{\enskip Work done as an intern at the Allen Institute for AI.}\quad Matt Gardner$^\clubsuit$\quad Hannaneh Hajishirzi$^{\diamondsuit\clubsuit}$ \\ \textbf{Tushar Khot$^\clubsuit$ \quad Pradeep Dasigi$^\clubsuit$}\\
  $^\diamondsuit$University of Washington\\
  $^\clubsuit$Allen Institute for AI\\
  \texttt{\{jfferg,hannaneh\}@cs.washington.edu}\\ \texttt{\{mattg,tushark,pradeepd\}@allenai.org}}

\date{}

\begin{document}
\maketitle
\begin{abstract}
Humans often have to read multiple documents to address their information needs. However, most existing reading comprehension (RC) tasks only focus on questions for which the contexts provide all the information required to answer them, thus not evaluating a system's performance at identifying a potential lack of sufficient information and locating sources for that information. To fill this gap, we present a dataset, \dataset, with more than 13K questions over paragraphs from English Wikipedia that provide only partial information to answer them, with the missing information occurring in one or more linked documents. The questions were written by crowd workers who did not have access to any of the linked documents, leading to questions that have little lexical overlap with the contexts where the answers appear. This process also gave many questions without answers, and those that require discrete reasoning, increasing the difficulty of the task. We follow recent modeling work on various reading comprehension datasets to construct a baseline model for this dataset, finding that it achieves 31.1\% F1 on this task, while estimated human performance is 88.4\%. The dataset, code for the baseline system, and
a leaderboard can be found
at \url{https://allennlp.org/iirc}.
\end{abstract}

\begin{figure}[t]
    \centering
    \includegraphics[width=\columnwidth]{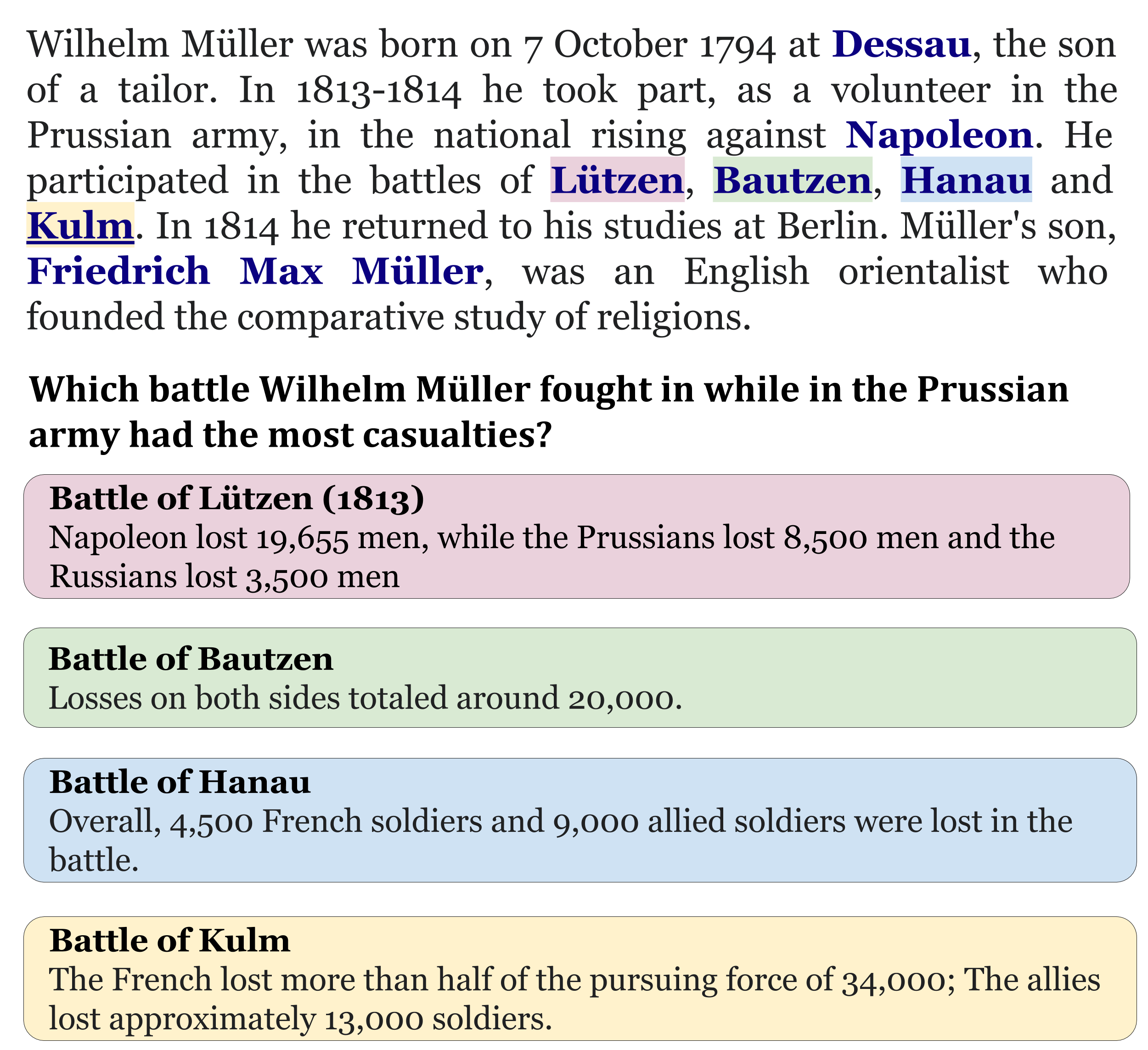}
    \caption{An example from \dataset{}. At the top is a context paragraph which provides only partial information required to answer the question. The bold spans in the context indicate links to other Wikipedia pages. The colored boxes below the question show snippets from four of these pages that provide the missing information for answering the question. The answer is the underlined span.}
    \label{fig:iirc_example}
\end{figure}
\section{Introduction}
Humans often read text with the goal of obtaining information. Given that a single document is unlikely to contain all the information a reader might need, the reading process frequently involves identifying the information present in the given document, and what is missing, followed by locating a different source that could potentially contain the missing information. Most recent reading comprehension tasks, such as SQuAD 2.0~\citep{rajpurkar2018know}, DROP~\citep{Dua2019DROPAR}, or Quoref~\citep{dasigi2019quoref}, evaluate models using a relatively simpler setup where all the information required to answer the questions (including judging them as being unanswerable) is provided in the associated contexts. While this setup has led to significant advances in reading comprehension~\citep{ran2019numnet,Zhang2020RetrospectiveRF}, the tasks are still limited since they do not evaluate the capability of models at identifying precisely what information, if any, is missing to answer a question, and where that information might be found.

On the other hand, open-domain question answering tasks \citep{Chen2017ReadingWT,joshi-etal-2017-triviaqa,Dhingra2017QuasarDF} present a model with a question by itself, requiring the model to retrieve relevant information from some corpus.  However, this approach loses grounding in a particular passage of text, and it has so far been challenging to collect diverse, complex question in this setting.

Alternatively, complex questions grounded in context can be converted to open-domain or incomplete-information QA datasets such as HotpotQA~\citep{yang2018hotpotqa}. However, they do not capture the information-seeking questions that arise from reading a single document with partial information \citep{Min2019CompositionalQD,Chen-2019-multihop-design}.

We present a new dataset of incomplete information reading comprehension questions, \dataset, to address both of these limitations. \dataset is a crowdsourced dataset of \numquestions{} questions over \numparagraphs{} paragraphs from English Wikipedia, with most of the questions requiring information from one or more documents hyperlinked to the associated paragraphs, in addition to the original paragraphs themselves. Our crowdsourcing process (Section~\ref{sec:dataset_creation}) ensures the questions are naturally information-seeking by decoupling question and answer collection pipelines. Crowd workers are instructed to ask follow-up questions after reading a paragraph, giving links to pages where they would expect to find the answer. This process results in questions like the one shown in Figure~\ref{fig:iirc_example}. As illustrated by the example, this setup results in questions requiring complex reasoning, with an estimated 39\% of the questions in \dataset requiring discrete reasoning. Moreover, $30\%$ of the questions in \dataset require more than one linked document in addition to the original paragraph and $30\%$ of them are unanswerable even given the additional information. When present, the answers are either extracted spans, boolean, or values resulting from numerical operations.

To evaluate the quality of the data, we run experiments with a modified version of NumNet+~\cite{ran2019numnet}, a state-of-the-art model from DROP~\cite{Dua2019DROPAR}, chosen because a significant portion of questions in \dataset require numerical reasoning similar to that found in DROP. Because DROP uses only a single paragraph of context, we add a two-stage pipeline to retrieve necessary context for the model from the linked articles. The pipeline first identifies which links are pertinent, and then selects the most relevant passage from each of those links, concatenating them to serve as input for the model (Section~\ref{sec:modeling}).

This baseline achieves an $F_1$ score of 31.1\% on \dataset, while the estimated human performance is 88.4\% $F_1$.  Even giving the model oracle pipeline components results in a performance of only 70.3\%. Taken together, these results show that substantial modeling is needed both to identify and retrieve missing information, and to combine the retrieved information to answer the question (Section~\ref{sec:experiments}). We additionally perform qualitative analysis of the data, and find that the errors of the baseline model are evenly split between retrieving incorrect information, identifying unanswerable questions, and successfully reasoning over the retrieved information.

By construction, all examples in \dataset require identifying missing information.  Even though current model performance is quite low, a model trained on this data could theoretically leverage that fact to achieve artificially high performance on test data, because it does not have to first determine \emph{whether} more information is needed. To account for this issue, we additionally sample questions from SQuAD 2.0~\citep{rajpurkar2018know} and DROP~\citep{Dua2019DROPAR}, which have similar question language to what is in \dataset, putting forward this kind of combined evaluation as a challenging benchmark for the community.  Predictably, our baseline model performs substantially worse in this setting, reaching only 28\% $F_1$ on the \dataset portion of this combined evaluation (Section~\ref{sec:combined}).

\begin{figure*}[t]
\centering
\includegraphics[width=\linewidth]{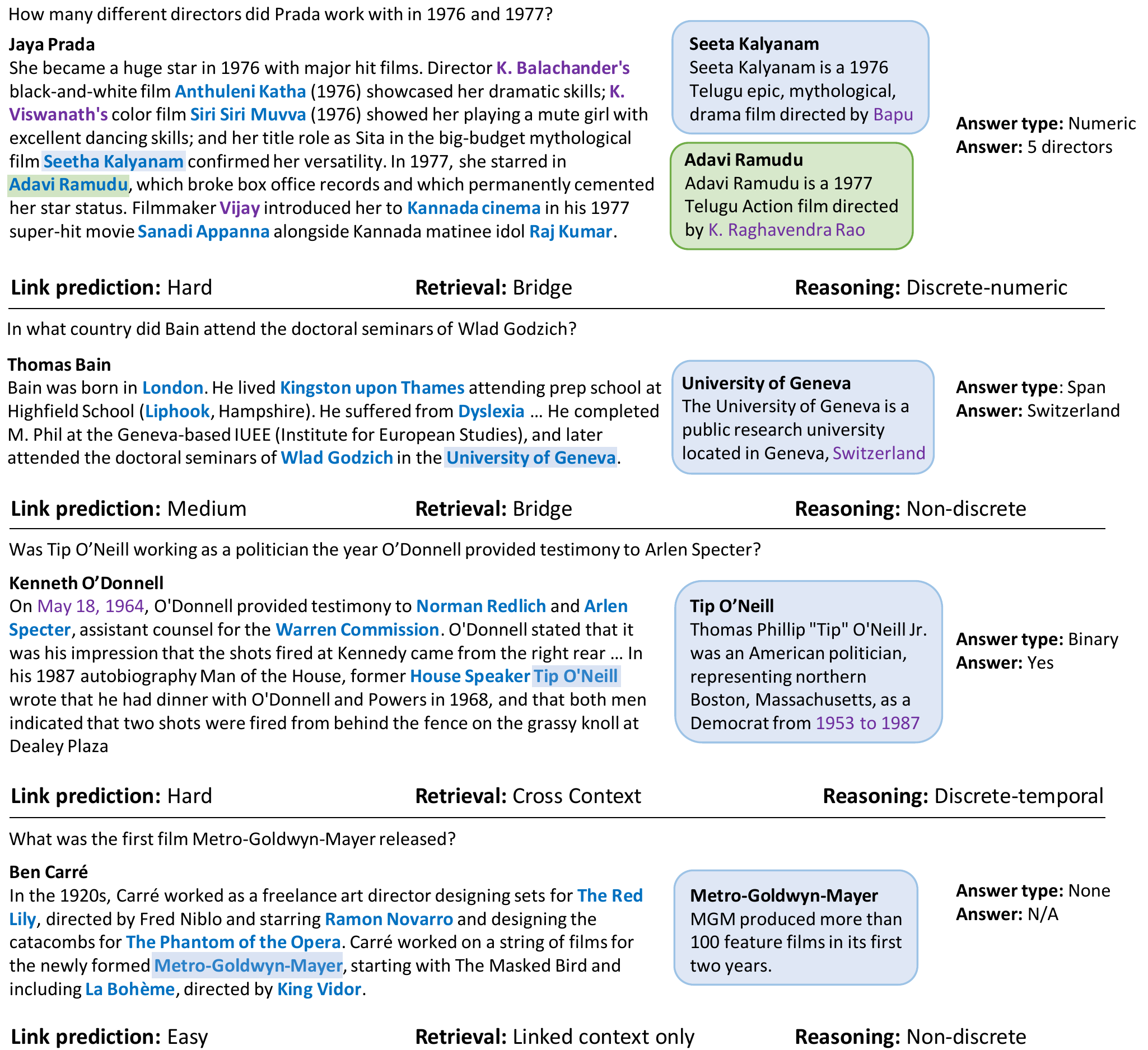}
\caption{Examples from \dataset, labeled with what kinds of processing are required to answer each question. See Table \ref{table:reasoning_types} for more details. The passages on the left are the original passage, with bold spans indicating links. The highlighted sections contain the necessary context found in linked articles. Purple highlights indicate either the answer, for the second question, or the information used to compute the answer.}
\label{fig:examples}

\end{figure*} 

\section{Building \dataset}\label{sec:dataset_creation}
We used Wikipedia to build \dataset and relied on the fact that entities in Wikipedia articles are linked to other articles about those entities, providing more information about them. Our goals were to build a dataset with naturally information-seeking questions anchored in paragraphs with incomplete information, such that identifying the location of missing information is non-trivial, and answering the questions would require complex cross-document reasoning. 

We ensured that the questions are information-seeking by separating question and answer collection processes, and by not providing the question writers access to the contexts where the answers occur. This process also ensured that we get questions that have minimal lexical overlap with the answer contexts. We used Wikipedia paragraphs with many outgoing links to increase the difficulty of identifying the articles that provide the missing information. To ensure complex cross-document reasoning, we asked the crowd workers to create questions that need information from the seed paragraph as well as one or more linked articles. This constraint resulted in questions that are answerable neither from the original paragraph alone, nor from one of the linked articles alone, often requiring over 3+ passages to answer. The remainder of this section describes our data collection process.

\begin{table*}[t]
\centering
\small
\begin{tabular}{llll}
\toprule
&\textbf{Type} & \textbf{Description} & \textbf{Percentage}\\
\midrule
 & Easy & Link is explicitly mentioned in the question & 41\%\\
Link Prediction & Medium & Context is required to determine link target & 47\%\\
& Hard & Context is required to determine link targets and number of links & 12\%\\
\midrule
& Linked context only & Original passage is not necessary to answer question & 14\%\\
Retrieval & Bridge & Original passage is only necessary to determine links & 57\%\\
& Cross context & Original passage is necessary to find relevant information in links & 29\%\\
\midrule
& Non-discrete & No discrete reasoning is required & 61\%\\
Reasoning & Discrete-numeric & Discrete reasoning is required & 11\%\\
& Discrete-temporal & Discrete reasoning involving time is required & 28\%\\
\midrule
\multirow{4}{*}{Answer} &
Span & Answer is one or more spans selected from question or context & 45\%\\
& Numeric & Answer is a number (with a unit provided) & 17\%\\
& Binary & Answer is either \textit{yes} or \textit{no} & 8\%\\
& None & Question cannot be answered given the provided context & 30\%\\
\bottomrule
\end{tabular}
\caption{Frequency of different types of retrieval, reasoning, and answers that appear in \dataset.}
\label{table:reasoning_types}
\end{table*}

\subsection{Seed Paragraphs}
We started by collecting paragraphs from Wikipedia articles containing ten or more links to other Wikipedia articles. This resulted in roughly $130$K passages. We then created two separate crowdsourcing tasks on Amazon Mechanical Turk\footnote{www.mturk.com}; one for collecting questions, and one for collecting answers. Workers for each task were chosen based on a qualification task. Their submissions were manually inspected, and those that produced high quality questions and correct answers, respectively, continued to work on the main annotation tasks. 

\subsection{Collecting Questions}
 Given a paragraph with links to other articles highlighted, crowd workers were tasked with writing questions that require information from the paragraph, as well as from one or more of linked articles. Workers could see the links, and the titles of the articles they pointed to, but not the contents of the linked articles. Since the linked articles were not provided, the workers were asked to frame questions based on the information they think would be contained in the those articles. For each human intelligence task (HIT), workers were presented with a collection of ten paragraphs, and were asked to write a total of ten questions using any of those paragraphs, with two questions requiring following two or more links. For example given a passage about an actor that mentions \textit{Rasulala had roles in Cool Breeze (1972), Blacula (1972), and Willie Dynamite (1973)}, an example of a question requiring multiple links would be \textit{How many different directors did Rasulala work with in 1972?}.
 
 In order to minimize questions with shortcut reasoning, we provided workers extensive instructions along with examples of good and bad questions to ask. Examples of bad questions included questions that did not require any links - \textit{Who did the Arakanese kings compare themselves to?} when the context included \textit{They compared themselves to Sultans}; and questions that did not require information from the original passage - \textit{What was Syed Alaol's most famous work?} when the context included \textit{Syed Alaol was a renowned poet}.
 
 In addition to writing questions, workers also provided the context from the original paragraph that they thought would be necessary to answer the question, as well as the links they expected to contain the remaining necessary information. Workers were paid $\$4.00$ per set of ten questions, and reported taking $25$ minutes on average, coming out to $\$9.60$ per hour. 40 workers passed the qualification and worked on the main task.

\subsection{Collecting Answers} 
For the answer task, workers were given a collection of ten questions, their respective original paragraphs, and the context/links selected by the question writer. For each paragraph, workers were able to see the links, and could follow them to view the text, not including tables or images, of the linked document. 

They were then asked to select an answer from one of four types: a span of text from either the question or a document, a number and unit, yes/no, or \textit{no answer}. For answerable questions, i.e. any of the first three types, they were additionally asked to provide the minimal context span(s), necessary to answer the question. For unanswerable questions, there is typically no indication that the answer is not given, so no such context can be provided. For example, the following question was written for a passage about a ship called the Italia: \textit{Who was the mayor of New York City when Italia was transferred to Genoa-NYC?} Following the link to New York City mentions the current mayor, but not past mayors, making it unanswerable.

Annotators were also given the option of labeling a question as bad if it didn't make sense, and these bad questions were then filtered out. For example, if an annotator misinterpreted the passage when writing the question as in the case of the following question written about a horse, Crystal Ocean, and St Leger, which the annotator thought was a horse, but is actually a horse race: \textit{Is Crystal Ocean taller than St Leger?}. Additionally, A small percentage of questions that can be answered from the original paragraph alone were also marked as being bad.

For the training set, comprising 80\% of the data, each question was answered by a single annotator. For the development and test sets, comprising 10\% each, three annotators answered each question, and only questions where at least two annotators agreed on the answer were kept. Workers were paid $\$3.00$ per set of ten answers, and reported taking $20$ minutes on average, coming out to $\$9.00$ per hour. 33 workers passed the qualification and worked on the main task.

\subsection{Dataset Analysis}
\label{sec:analysis}

In Figure \ref{fig:examples} we show some examples from \dataset, labeled with different kinds of processing required to solve them. The types are described in detail in Table \ref{table:reasoning_types}. These types and percentages were computed from a manual analysis of 100 examples. 

In Table \ref{table:data_statistics} we provide some global statistics of the dataset. In total, there are \numquestions{} questions over \numparagraphs{} passages. Each passage contains an average of 14.5 outgoing links. Using the context provided by the answer annotators, we are able to compute a distribution of the number of links required to answer questions in the dataset, included in Table \ref{table:num_links_breakdown}. While the majority of questions require information from only one linked document in addition to the original paragraph, $30\%$ of questions require two or more, with some requiring reasoning over as many as $12$ documents to reach the answer. This variability in the number of context documents adds an extra layer of complexity to the task.

We also analyzed the initial trigrams of questions to quantify the diversity of questions in the dataset. We found that the most common type of questions, those related to time (eg “How old was”, “How long did”), make up 15\% of questions. There are 3.5k different initial trigrams across the 10.8k questions in the training set.

\begin{table}[t]
\centering
\small
\begin{tabular}{ll}
\toprule
Number of questions & \numquestions\\
Number of passages & 5698\\
Average number of links per passage & 14.5\\
Average passage length (words) & 197.5\\ 
Average question length (words) & 13.6\\
\bottomrule
\end{tabular}
\caption{Statistics of \dataset.}
\label{table:data_statistics}
\end{table}

\section{Modeling \dataset}
\label{sec:modeling}
\subsection{Task Overview}
Formally, a system tackling \dataset is provided with the following inputs: a question $Q$; a passage $P$ ; a set of links contained in the passage, $L = \{l_i\}_{i=1}^N$; and the set of articles those links lead to, $\textbf{A} = \{a^i\}_{i=1}^N$. The surface form of each link, $l_i$ is a sequence of tokens in $P$ and is linked to an article $a^i$. The target output is either a number, a sequence of tokens in one of $P$, $Q$, or $a^i$, \texttt{Yes}, \texttt{No}, or \texttt{NULL} (for unanswerable questions).

\subsection{Baseline Model}\label{sec:baselines}
To evaluate the difficulty of \dataset, we construct a baseline model adapted from a state-of-the-art model built for DROP. We choose a DROP model due to the inclusion of numerical reasoning questions in our dataset. Because the model was not originally used for data requiring multiple paragraphs and retrieval, we first predict relevant context to serve as input to the QA model using a pipeline with three stages: 
\begin{enumerate}[noitemsep]
    \item Identify relevant links
    \item Select passages from linked articles
    \item Pass the concatenated passages to a QA model
\end{enumerate}

\subsubsection{Identifying Links}
To identify the set of relevant links, $\text{L}^\prime$, in a passage, P, for a question, Q, the model first encodes the concatenation of the question and original passage using BERT~\citep{devlin2019bert}. It then concatenates the encoded representations of the first and last tokens of each link as input to a scoring function, following the span classification procedure used by \citet{joshi-etal-2019-spanbert}, selecting any links that score above a threshold $g$.
 \begin{flalign*}
 P^\prime &= \text{BERT}([Q || P]) \\
 \text{Score}(\text{l}) &= f([p^\prime_i \| p^\prime_j]), \hspace{.2cm}l=(p_i ... p_j, a) \\
 L^\prime &= \{l : \text{Score}(l) > g\}
 \end{flalign*}
 where $l$ is a link covering tokens $p_i...p_j$ linking to article $a$. 
 
\subsubsection{Selecting Context}
Given the set, $L'$ from the previous step, the model then must select relevant context passages from the documents. For each document, it first splits the document into overlapping windows\footnote{See section \ref{sec:implementation} for more details.}, $w_0, w_1...w_n$. Each window is then concatenated with the question and prepended with a \texttt{CLS} token, and encoded with BERT. The encoded \texttt{CLS} tokens are then passed through a linear predictor to score each window, and the highest scoring sections from each document are concatenated as context for the final model, $C$. 
\begin{flalign*}
    c_{a_i} &= \max_{w_j \in \text{Split}(a_i)} f(\text{BERT}([Q||w_j])) \\
    C &= [c_{a_i}: a_i \in L']
\end{flalign*}

\subsubsection{QA Model}
As mentioned above, the final step in the pipeline is passing the concatenated context, along with the question and a selected window from the original passage, as input to a QA model. For our experiments, we use NumNet+, because it is the best performing model on the DROP leaderboard with publicly available code. At a high level, NumNet+ encodes the input using RoBERTa~\citep{Liu2019RoBERTaAR}, as well as a numerical reasoning component. It then passes these into a classifier to determine the type of answer expected by the question, which we modified by adding binary and unanswerable as additional answer types. This model is trained using the gold context for answerable questions, and predicted context for unanswerable questions. We do this because by definition, unanswerable questions do not have annotated answer context. 

\section{Experiments}
\label{sec:experiments}
\subsection{Evaluation Metrics}
We use two evaluation metrics to compare model performance: Exact-Match (EM), and a numeracy-focused (macro-averaged) F1 score, which measures overlap between a bag-of-words representation of the gold and predicted answers. Due to the number of numeric answers in the data, we follow the evaluation methods used by DROP~\cite{Dua2019DROPAR}. 

Specifically, we employ the same implementation of Exact-Match accuracy as used by SQuAD~\cite{rajpurkar-etal-2016-squad}, which removes articles and does other simple normalization, and our F1 score is based on that used by SQuAD. We define F1 to be 0 when there is a number mismatch between the gold and predicted answers, regardless of other word overlap. When an answer has multiple spans, we first perform a one-to-one alignment greedily based on bag-of-word overlap on the set of spans and then compute average F1 over each span. For numeric answers, we ignore the units. Binary and unanswerable questions are both treated as span questions. In the unanswerable case, the answer is a special \texttt{NONE} token, and in the binary case, the answer is either \textit{yes} or \textit{no}. 

\subsection{Implementation Details}
\label{sec:implementation}
For the link selection model, we initialized the encoder with pretrained BERT-base, and fine-tuned it during training. For the scoring function, we used a single linear layer with a sigmoid activation function. The model was trained using Adam, and the score threshold to select links was set to $0.5$. Additionally, we truncated any passages longer than $512$ tokens to $512$. This occurred in less than $1\%$ of the data. This model is trained using a cross-entropy objective with the information provided in the gold context by annotators. Any links pointing to articles with an annotated context span are labeled 1, and all other links are labeled 0. 

For the passage selection model, we again initialized the encoder with pretrained BERT-base, and fine-tuned it during training. We set the window size such that the concatenation of all selected contexts, along with the question and a selection from the original passage, has max length $512$. More specifically, using the number of links, $N_l$ selected in the previous step, for a question with $N_Q$ tokens, we set the window size to be $\frac{512-(N_Q)}{N_l+1}$. We set the stride to be $\frac{1}{4}$ the window size, i.e. if the first window contains tokens $[0,200]$, the second window would contain $[50,250]$. We used a single linear layer with a sigmoid activation as the scoring function. We train this model with a cross-entropy objective. We use the gold context provided by annotators, labeling sections that contain the entirety of the annotated context 1, and all other sections 0. 

For NumNet+, we followed the hyperparameter and training settings specified in the original paper \citep{ran2019numnet}. We trained the model on gold context provided by annotators when available, i.e. for answerable questions, and predicted context from the previous steps otherwise.

\subsection{Results and Discussion}\label{sec:results}
\begin{table}
\small
\centering
\begin{tabular}{lrrrr}
\toprule
&\multicolumn{2}{c}{Dev} &\multicolumn{2}{c}{Test}\\
\textbf{Model} & \textbf{EM} & \textbf{F1} & \textbf{EM} & \textbf{F1}\\
\midrule
Full model & 29.6 & 33.0 & 27.7 & 31.1\\
Oracle L & 30.9 & 34.7 & 29.0 & 32.5\\
Oracle L+C & 63.9 & 69.2 & 65.6 & 70.3 \\
Human  & - & - & 85.7 & 88.4 \\
\bottomrule
\end{tabular}
\caption{Baseline and oracle results on \dataset. Human evaluation was obtained from a subset of 200 examples from the test set. We evaluate the model when given oracle links (L) and retrieved contexts (C). Retrieving the correct contexts is a significant challenge, but even given oracle contexts there is a substantial gap between model and human performance.}
\label{table:answer_distribution}
\end{table}

\paragraph{Full Task Results} Table \ref{table:answer_distribution} presents the performance of the baseline model. It additionally shows the results of using gold information at each stage of the pipeline, as well as human performance on computed on a subset of 200 examples from the test set. 
The model achieves 31.1\% $F_1$, which is well below the human performance of 88.4\%. Even with the benefit of the gold input, there is still room for improvement on reasoning over multiple contexts, as performance is still 18\% absolute below human levels. 
The model does a good job of predicting the relevant links, as evidenced by the fact that using the gold links only improves performance by 1 point, but still struggles to identify the appropriate context within the linked documents. This is likely due to annotators not being able to see the linked context when the questions are written. This makes this step more difficult by not providing the model with surface-level lexical cues in the question that it could use to easily select the appropriate context.

\begin{table}
\small
\centering
\begin{tabular}{lcc}
\toprule
\textbf{Number of links} & \textbf{EM} & \textbf{F1}\\
\midrule
1 (70\%)&33.2&36.7\\
2 (23\%)&27.0&30.6\\
3 (4\%)&25.9&31.5\\
4+ (3\%)&40.9&43.4\\
\bottomrule
\end{tabular}
\caption{Exact match and F1 of the baseline model on the \dataset dev set broken down by number of links necessary to answer the question. The numbers in parentheses are the percentage of questions in the full dataset that require that number of context documents.}
\label{table:num_links_breakdown}
\end{table}

\paragraph{Analysis of Number of Linked Documents} Table \ref{table:num_links_breakdown} shows the results of running the full pipeline broken down according to the number of linked documents required to answer the question. These performance  differences are the result of a few factors. The first is the fact that the more links required to answer a question, the more chances there are for failure to retrieve the necessary information. This is exacerbated by the pipeline nature of our baseline model. However, the spike in performance for questions requiring four or more links is caused by the number of unanswerable questions. Nearly half of the questions in that category are unanswerable, and the model largely predicts \textit{No Answer} on those questions. Finally, the distribution of question types is different conditioned on the number of links. Questions that require more links often also require some form of discrete reasoning, which is more difficult for the model to handle.

\begin{table}
\small
\centering
\begin{tabular}{lrr}
\toprule
\textbf{Answer Type} & \textbf{EM} & \textbf{F1}\\
\midrule
Span & 24.0 & 29.1\\
Number & 20.4 & -\\
Binary & 56.5 & -\\
No Answer & 32.4 & -\\
\bottomrule
\end{tabular}
\caption{Exact match and F1 of the baseline model on the \dataset dev set broken down by answer type. F1 equals EM for non-span types, so is not repeated.}
\label{table:answer_type_breakdown}
\end{table}

\paragraph{Analyzing Different Answer Types} Table \ref{table:answer_type_breakdown}  shows the performance broken down according to the type of answer each question has. The model performs worst on questions with numeric answers. This is due to the fact that these questions often require the model to do arithmetic to solve, which, as discussed above, the model struggles with relative to other types of questions. 

\begin{table}
\small
\centering
\begin{tabular}{lrrr}
\toprule
\textbf{Input} & \textbf{P} & \textbf{R} & \textbf{F1} \\
\midrule
Constant baseline & 26.7 & 100.0 & 42.1\\
Question  only & 61.8 & 54.9 & 58.1\\
Question + Passage & 64.2 & 54.9 & 59.2\\
Question + Pred Context & 62.3 & 70.1 & 66.0\\
\bottomrule
\end{tabular}
\caption{Precision, recall, and F1 of identifying unanswerable questions in the dev set with various baselines that use different combinations of the question, original passage, and predicted context.}
\label{table:no_answer_prediction}
\end{table}

\paragraph{Unanswerable Questions} Table \ref{table:no_answer_prediction} shows how well a simple model can identify unanswerable questions with varying amounts of information. We set this up as a binary prediction, either answerable or not, and use a linear classifier that takes the BERT \texttt{CLS} token as input. We also include the result of always predicting unanswerable as a baseline. When the model can only see the question, it improves over the baseline by around $10$ F1, meaning that there is some signal in the question alone, without any context. 

Some types of questions are more likely to be unanswerable, such as those asking for information with regards to a specific year, i.e. \textit{What was the population of New York in 1989?}. This is caused by Wikipedia more generally including current statistics, but not including a specific information for all previous years.
Additionally adding the original passage does not significantly improve performance. This is not surprising, as the original passage always contains information relevant to the question, and the question annotators could see that text when writing the question.

\subsection{Error Analysis}
In order to better understand the challenges of the dataset, we manually analyzed 100 erroneous predictions by the model.

\paragraph{Incorrect context (39\%)} These are the cases where the model identified the correct links but selected the wrong portion of the linked document. It often selects semantically similar context but misses the crucial information, e.g. selecting the duration instead of end date.

\paragraph{Modeling errors (32\%)} These are the cases in which the context passed to the final QA model contained all of the necessary information, but the model failed to predict the correct answer. This occurred most commonly for questions requiring math, with the model including unnecessary dates in the computation, resulting in predictions that were orders of magnitude off. For example, predicting \textit{-1984} when the question was asking for the age of a person.

\paragraph{Identifying unanswerable questions (24\%)} In these cases, the QA model was provided with related context that was missing crucial information, similar to the first class of errors. However, in this case, the full articles also did not contain the necessary information. In these cases the model often selected a related entity, ie for a question asking \textit{In which ocean is the island nation located?}, the model predicted the island nation, \textit{Papua New Guinea} as opposed to the ocean, which was not mentioned.

\paragraph{Insufficient Links (5\%)} These are cases where insufficient links were selected from the original passage, thus not providing enough information to answer the question. While the model can handle over-selection of links, we found that the vast majority of the time, the system correctly identified both the necessary and sufficient links, rarely over-predicting the required links.

\section{Combined Evaluation}\label{sec:combined}
By construction, all the questions in \dataset require more than the original paragraph to answer.  This means that a reading comprehension model built for \dataset does not actually have to detect \emph{whether} more information is required than what is in the given paragraph, as it can always assume that this is true.  In order to combat this bias, we recommend an additional, more stringent evaluation that combines \dataset with other reading comprehension datasets that do not require retrieving additional information.  This is in line with recently-recommended evaluation methodologies for reading comprehension models~\cite{talmor-berant-2019-multiqa,dua-etal-2019-comprehensive}.

In this section, we present the results of one such evaluation.  Noting that \dataset has similar properties to both SQuAD 2.0~\cite{rajpurkar2018know} and DROP~\cite{Dua2019DROPAR}, and even similar question language in places, we sample questions from these datasets to form a combined dataset for training and evaluating our baseline model.

\paragraph{Sampling from SQuAD 2.0 and DROP}
To construct the data for the combined evaluation, we sample an additional $3360$ questions from SQuAD 2.0 and DROP, so that they make up $20\%$ of the questions in the new data. We sample from SQuAD 2.0 and DROP with a ratio of $3:1$ in order to match the distribution of numeric questions in \dataset and used a Wikifier~\citep{ChengRo13} to identify the links to Wikipedia articles in them.

\begin{table}
    \small
    \centering
    \begin{tabular}{llllll}
         \toprule
         \textbf{Training} & \multicolumn{3}{c}{\textbf{Links}} & \multicolumn{2}{c}{\textbf{QA}} \\
         \textbf{Data} & \textbf{P} & \textbf{R} & \textbf{F1} & \textbf{EM} & \textbf{F1}\\
         \midrule
         IIRC & 88 & 98 & 93 & 32.0& 35.6 \\ 
         IIRC + S + D& 85 & 79 &82 & 24.6& 28.0 \\
         \bottomrule
    \end{tabular}
    \caption{Results for link identification and QA when training the baseline model on \dataset and sampled questions from SQuAD (S) and DROP (D). }
    \label{table:combination_results}
\end{table}

\paragraph{Results}
We train the full baseline on \dataset augmented with sampled DROP and SQuAD data, and evaluate it on the \dataset dev set without any additional sampled data. We don't include any sampled data in the evaluation in order to make a direct comparison to \dataset to see how adding questions that don't require external context affects the model's ability to identify necessary context. We also include the results of running just the link identification model trained under each setting. We show the results in table \ref{table:combination_results}. Adding the extra dimension of determining whether extra information is necessary causes the model to become less confident, significantly hurting recall on link selection. These missed predictions then propagate down the pipeline, resulting in a loss of almost 8\% $F_1$ when compared to a model trained on just \dataset.

We also evaluated the combination model on a dev set with sampled SQuAD and DROP data to see how well the model learned to identify that no external information was necessary. Given that none of the SQuAD or DROP data requires external links, this evaluation could only negatively impact precision. We find that precision dropped by 8 points, compared to a drop of 28 points when the model trained only on \dataset was used, indicating that the model is able to learn to identify when no external information is required.

\section{Related Work}
\paragraph{Questions requiring multiple contexts} Prior multi-context reading comprehension datasets were built by starting from discontiguous contexts, and forming compositional questions by stringing multiple facts either by relying on knowledge graphs as in QAngaroo~\citep{welbl2018constructing}, or by having crowdworkers do so, as in HotpotQA~\citep{yang2018hotpotqa}. It has been shown that many of these questions can be answered by focusing on just one of the facts used for building the questions~\citep{Min2019CompositionalQD}. In contrast, each question in \dataset was written by a crowdworker who had access to just one paragraph, with the goal of obtaining information missing in it, thus minimizing lexical overlap between questions and the answer contexts. Additionally, \dataset provides a unique question type: questions requiring aggregating information from many related documents, such as the second question in Figure \ref{fig:examples}. %As we show in Section~\ref{sec:results}, this process results in questions that cannot be answered from a single context.

\paragraph{Separation of questions from answer contexts} Many prior datasets (e.g.: WhoDidWhat~\citep{Onishi2016WhoDW}, NewsQA~\citep{trischler-etal-2016-newsqa}, DuoRC~\citep{Saha2018DuoRCTC}, Natural Questions~\citep{Kwiatkowski2019NaturalQA}, TyDiQA~\citep{Clark2020TyDiQA}) have tried to remove simple lexical heuristics from reading comprehension tasks by separating the contexts that questions are anchored in from those that are used to answer them. \dataset also separates the two contexts, but is unique given that the linked documents elaborate on the information present in the original contexts, naturally giving rise to follow-up questions, instead of open-ended ones. 

\paragraph{Open-domain question answering} In the open-domain QA setting, a system is given a question without any associated context, and must retrieve the necessary context to answer the question~\cite{Chen2017ReadingWT,joshi-etal-2017-triviaqa,Dhingra2017QuasarDF,yang2018hotpotqa,Seo-Open,karpukhin2020dense,min2019discrete}. \dataset is similar in that it also requires the retrieval of missing information. However, the questions are grounded in a given paragraph, meaning that a system must examine more than just the question in order to know what to retrieve.  Most questions in \dataset do not make sense in an open-domain setting, without their associated paragraphs.

\paragraph{Unanswerable questions} Unlike SQuAD 2.0~\citep{rajpurkar2018know} where the unanswerable questions were written to be close to answerable questions, \dataset contains naturally unanswerable questions that were not written with the goal of being unanswerable, a property that our dataset shares with NewsQA~\cite{trischler-etal-2016-newsqa}, Natural Questions~\cite{Kwiatkowski2019NaturalQA}, and TyDi QA~\cite{Clark2020TyDiQA}. Results shown in Section~\ref{sec:results} indicate that these questions cannot be trivially distinguished from answerable questions.

\paragraph{Incomplete Information QA} A few prior datasets have explored question answering given incomplete information, such as science facts~\cite{openbookqa,whatsmissing}. However, these datasets contain multiple choice questions, and the answer choices provide hints as to what information may be needed. \citet{Yuan-2020-imrc} explore this as well using a POMDP in which the context in existing QA datasets is hidden from the model until it explicitly searches for it.
% Additionally, it is unclear if the knowledge needed to answer the questions is even present in any knowledge base.

\section{Conclusion}

We introduced \dataset, a new dataset of incomplete-information reading comprehension questions.  These questions require identifying what information is missing from a paragraph in order to answer a question, predicting where to find it, then synthesizing the retrieved information in complex ways.  Our baseline model, built on top of state-of-the-art models for the most closely related existing datasets, performs quite poorly in this setting, even when given oracle retrieval results, and especially when combined with other reading comprehension datasets.  
\dataset both provides a promising new avenue for studying complex reading and retrieval problems and demonstrates that much more research is needed in this area.

\subsection*{Acknowledgements}
We would like to acknowledge grants from ONR N00014-18-1-2826 and  DARPA N66001-19-2-403, and gifts from the Sloan Foundation and Allen Institute for AI. Authors would also like to thank members of the Allen Institute for AI, UW-NLP, and the H2Lab at The University of Washington for their valuable feedback and comments.
\bibliographystyle{acl_natbib}
\bibliography{emnlp2020}

\end{document}